\documentclass[fleqn,10pt]{wlscirep}
\usepackage[utf8]{inputenc}
\usepackage[T1]{fontenc}
\usepackage{subfigure}
\title{RetiNerveNet: Using Recursive Deep Learning to Estimate Pointwise 24-2 Visual Field Data based on Retinal Structure}

\author[1]{Shounak Datta}
\author[2]{Eduardo B. Mariottoni}
\author[1]{David Dov}
\author[2]{Alessandro A. Jammal}
\author[1]{Lawrence Carin}
\author[1,2,*]{Felipe A. Medeiros}
\affil[1]{Department of Electrical and Computer Engineering, Pratt School of Engineering, Duke University, Durham, NC, 27708, USA}
\affil[2]{Vision, Imaging and Performance (VIP) Laboratory, Duke Eye Center, Duke University, Durham, NC, 27705, USA}
\affil[*]{felipe.medeiros@duke.edu}

\begin{abstract}
Glaucoma is the leading cause of irreversible blindness in the world, affecting over 70 million people. The cumbersome Standard Automated Perimetry (SAP) test is most frequently used to detect visual loss due to glaucoma. Due to the SAP test's innate difficulty and its high test-retest variability, we propose the RetiNerveNet, a deep convolutional recursive neural network for obtaining estimates of the SAP visual field. RetiNerveNet uses information from the more objective Spectral-Domain Optical Coherence Tomography (SDOCT). RetiNerveNet attempts to trace-back the arcuate convergence of the retinal nerve fibers, starting from the Retinal Nerve Fiber Layer (RNFL) thickness around the optic disc, to estimate individual age-corrected 24-2 SAP values. Recursive passes through the proposed network sequentially yield estimates of the visual locations progressively farther from the optic disc. While all the methods used for our experiments exhibit lower performance for the advanced disease group, the proposed network is observed to be more accurate than all the baselines for estimating the individual visual field values. We further augment RetiNerveNet to additionally predict the SAP Mean Deviation values and also create an ensemble of RetiNerveNets that further improves the performance, by increasingly weighting-up underrepresented parts of the training data.
\end{abstract}
\begin{document}

\flushbottom
\maketitle
\thispagestyle{empty}

\section*{Introduction}

Glaucoma is the leading cause of irreversible blindness in the world, and it is estimated that over 70 million people are affected by it \cite{quigley2006number}. Defined as a chronic and progressive optic neuropathy, it is characterized by the progressive loss of retinal ganglion cells, resulting in characteristic changes in the optic disc and consequent visual loss \cite{weinreb2014pathophysiology}. Although irreversible, adequate treatment can slow or even prevent further damage caused by the disease \cite{garway2015latanoprost}. This highlights the need for early diagnosis and accurate prediction of glaucoma progression. In clinical practice, investigation and monitoring of patients with glaucoma involves evaluation of the visual field using Standard Automated Perimetry (SAP), and evaluation of the optic disc and Retinal Nerve Fiber Layer (RNFL) using Optical Coherence Tomography (OCT). 

Currently, SAP is the most frequently used method for detecting visual loss due to glaucoma \cite{prum2016primary}. It is most commonly performed using the $24-2$ Swedish Interactive Threshold Algorithm \cite{bengtsson1997new} of the Humphrey perimeter (Carl-Zeiss Meditec, Inc.), by presenting luminous stimuli with varying intensities at 54 different locations of the visual field and recording the patient's response to the stimuli. The patient is typically required to press a switch each time they are able to see a presented stimulus. The dimmest stimulus detected by the patient in each visual field location is recorded using a logarithmic scale, where higher values represent lower brightness, and therefore better results. Although SAP is the mainstay strategy for visual field testing, there are issues associated with the testing strategy. Due to the interactive nature of the test, there is typically a learning curve for the patient to comprehend the test instructions. Therefore, for many patients, the first few tests cannot be relied on. Further, a high level of attention is demanded of the patient during the entire duration of the test. Fatigue and lack of attention often leads to artifacts and to unreliable results. Finally, even for well-trained patients, there is high test-retest variability associated with SAP, which can preclude or delay the diagnosis or detection of disease progression \cite{gracitelli2018detection, urata2020comparison}.

With the advent of a family of better ocular imaging technologies, namely the Spectral-Domain OCT (SDOCT), it has become possible to image the retina with near-histologic definition, with an axial resolution of a few micrometers. Using the SDOCT peripapillary scan, it is possible to measure the thickness of the RNFL at evenly spaced points on a circle centered around the opening of the optic disc. In contrast to SAP, SDOCT is fast, highly reproducible, and does not require any response from the patient. However, while SDOCT is capable of accurately measuring structural changes in the retina and the optic disc, it is not a direct assessment of visual function.

Previous works have shown that it is indeed possible to draw conclusions about the visual function based on the structural information acquired with SDOCT \cite{zhu2010predicting, zhu2011quantifying}. Such estimates can potentially be used as proxies for visual function, particularly for patients who are incapable of undergoing SAP. Deep learning techniques, Convolutional Neural Networks (CNNs) in particular \cite{lecun1998gradient}, are capable of using spatial information to identify underlying relationships that may not be easily discerned by conventional methods. Some existing studies \cite{maetschke2019inference, christopher2020deep}, have used such deep learning techniques to estimate SAP summary metrics like Mean Deviation (MD, a weighted average of the age-corrected visual field values) using information acquired with SDOCT. Other works\cite{guo2017optical, mariottoni2020artificial, park2020deep}, more closely related to our own, have attempted to estimate pointwise sensitivities for all the locations tested by SAP, based on SDOCT measurements. However, these previous investigations have not made use of the known topographic characteristics of the RNFL when attempting to estimate SAP data \cite{garway2000mapping}.

\begin{figure}[ht]
    \centering
        \subfigure[]{%
            \label{fig:rnfl1}% label for this sub-figure
            \includegraphics[width=0.25\linewidth]{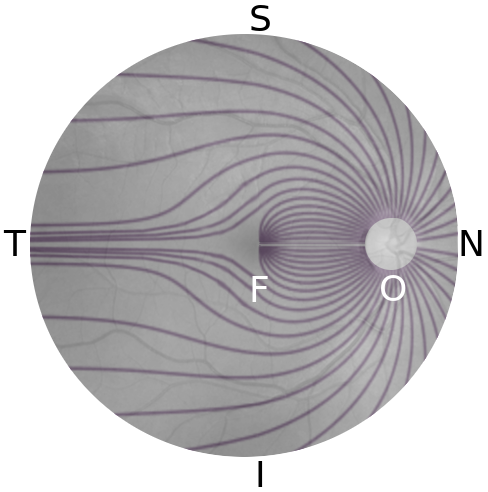}
        }%\qquad % if you want to space out the images a bit
        \subfigure[]{%
            \label{fig:rnfl2}% label for this sub-figure
            \includegraphics[width=0.25\linewidth]{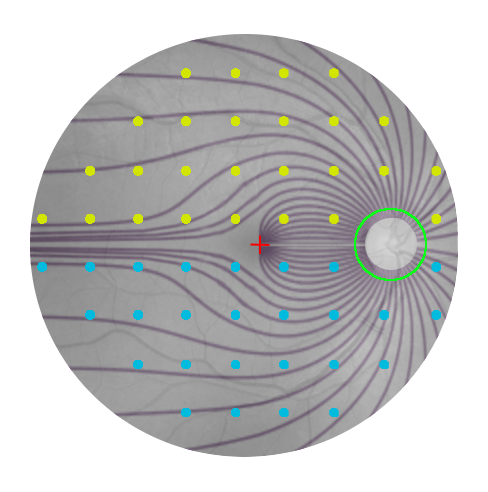}
        }
    \caption{Structure of the RNFL and its relationship with the SDOCT and SAP tests (best viewed in color): (a) The axons from the retinal ganglion cells converge to the optic disc in a characteristic arcuate pattern while deviating from the fovea (F) \cite{airaksinen2008conformal}. The temporal, superior, nasal and inferior regions of the retina are respectively labeled T, S, N, and I while the optic disc is labeled O. (b) SAP measures visual function at 54 locations (2 location near the blind spot being excluded in our analysis) on the retina (cyan and yellow dots), centered (red `+') at the fovea. Because of lateral inversion in the eye, the SAP locations in the superior hemiretina (yellow dots) are responsible for inferior vision while those in the inferior hemiretina (cyan dots) are responsible for superior vision. SDOCT measures the RNFL thickness at 768 equally-spaced locations along the (green) circle around the optic disc.}\label{fig:rnfl}%  We intend to use the RNFL thickness at 768 equally-spaced locations on this circle to estimate the visual function values at the 52 SAP locations we consider.
\end{figure}

The RNFL is comprised of the axons of the retinal ganglion cells which are responsible for carrying visual stimuli to the optic nerve. The ganglion cells converge to the optic disc in a characteristic arcuate pattern, deviating radially from the fovea, without crossing into the opposite half of the retina (see Figure \ref{fig:rnfl}), eventually emerging from the eye as the optic nerve \cite{weinreb2004primary}. Since our aim is to estimate visual function at different points on the retina based on the RNFL thicknesses around the optic disc, we must essentially trace-back the path along which the retinal axon fibers converge to the optic disc. We hypothesize that this approach will allow us to relate the amount of nerve tissue around the optic disc (represented by the RNFL thickness) to the appropriate SAP locations likely to be affected by it, thus improving the estimates. To this end, we propose RetiNerveNet (a.k.a. RetiNN), a deep fully convolutional neural network architecture that is inspired by this structure of retinal axon fibers. To the best of our knowledge, RetiNerveNet is the first method that incorporates elements of the structure of the retinal ganglion axon fibers for the task of estimating all the individual visual field values based on SDOCT information. Additionally, we augment the proposed deep neural architecture to simultaneously predict the individual age-corrected SAP values as well as the MD values and show that an ensemble of such networks can be employed to further improve performance.

\section*{Results}

The proposed RetiNerveNet is a deep fully convolutional neural network consisting of two similarly structured and separate sub-networks modeling the inferior and superior hemiretinae. Each sub-network, in turn, consists of four main blocks. The first convolutional block is meant to extract coarse structural information from the fine-grained RNFL thickness values. The second block is a Recursive Progression Layer (RPL, see Methods). Each recursive pass through the RPL is meant to model further spatial progression through the retina outwards from the optic disc. The third block extracts rich information about the outputs from the RPL by projecting onto different subspaces, while the fourth combines the different representations obtained from the former block into a single scalar output. Hence, recursive passes through RetiNerveNet sequentially yield estimates of the visual functions for locations on the retina which are progressively farther from the optic disc. 

\begin{figure}[ht]
    \centering
        \subfigure[]{%
            \label{fig:hist1}% label for this sub-figure
            \includegraphics[width=0.33\linewidth]{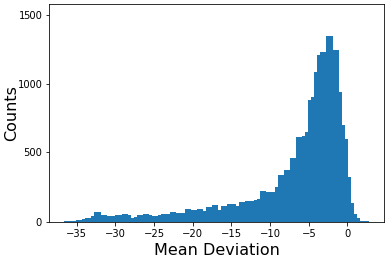}
        }%\qquad % if you want to space out the images a bit
        \subfigure[]{%
            \label{fig:hist2}% label for this sub-figure
            \includegraphics[width=0.33\linewidth]{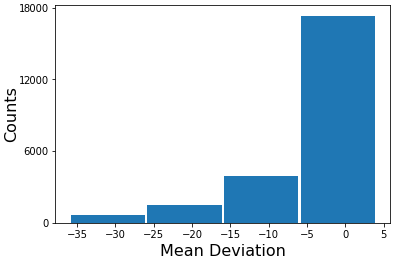}
        }
    \caption{Distribution of the training data: (a) The data is characterized by an uneven distribution across different MD values, with a long tail towards the low MD values. (b) The bulk of the data has MD values better than -6dB. The amount of data progressively diminishes through the other intervals.}\label{fig:hist}
\end{figure}

Our retrospective study is based on cross-sectional data about paired SDOCT and SAP tests, conducted within 180 days of each other. The data consisted of 38434 pairs of tests on 23171 eyes of 13284 patients. 1827 of these were from 463 eyes of 235 healthy individuals, the rest being from patients having or suspected of having glaucoma. The data was split into training, validation and testing sets, respectively consisting of 23060, 7687, and 7687 pairs of SDOCT-SAP tests, with all tests for a given patient always assigned to the same set. The training, validation as well as test data are characterized by an uneven distribution across MD values, with a long tail for the low MD values (see Figure \ref{fig:hist}). The demographic and clinical characteristics of the data are detailed in Table \ref{tbl:demographics}.

\begin{table}[ht]
%   \begin{center}
%   \begin{scriptsize}
    \centering
      \begin{tabular}{|l|l|l|l|l|}
        \hline
        Characteristic            & Early/No disease & Moderate disease & Advanced disease & Overall \\
        \hline
        \# SDOCT-SAP pairs        & 27216          & 5387           & 5831           & 38434 \\
        \hline
        \# SDOCT                  & 26860          & 5289           & 5752           & 37901 \\
        \hline
        \# SAP                    & 24055          & 4603           & 4920           & 33578 \\
        \hline
        \# Eyes                   & 16856          & 3638           & 3885           & 23171 \\
        \hline
        \# Patients               & 10440          & 3123           & 3062           & 13284 \\
        \hline
        Mean Age (SD)             & 54.9 (20.4)    & 59.1 (21.2)    & 59.7 (20.7)    & 56.2 (20.7) \\
        \hline
        \# Female (\%)            & 6039 (57.8\%)  & 1805 (57.8\%)  & 1585 (51.8\%)  & 9429 (56.7\%) \\
        \hline
        \# Race (\%)              &                &                &                & \\
        Black                     & 2831 (27.1\%)  & 959 (30.7\%)   & 1156 (37.8\%)  & 4946 (29.8\%) \\
        White                     & 6407 (61.4\%)  & 1826 (58.5\%)  & 1540 (50.3\%)  & 9773 (58.8\%) \\
        Other                     & 1202 (11.5\%)  & 338 (10.8\%)   & 366 (12.0\%)   & 1888 (11.4\%) \\
        \hline
        Mean RNFL Thickness       &                &                &                & \\
        in $\mu$m (SD)            & 87.5 (17.5)    & 73.6 (20.9)    & 64.2 (23.4)    & 82.0 (21.0) \\
        \hline
        Median MD in dB (IQR)     & -1.6 (-3.1, -0.3)               & -8.2 (-9.8, -6.9)               & -18.8 (-25.1, -14.9)              & -2.7 (-6.9, -0.8) \\
        \hline
        Median PSD in dB (IQR)    & 1.9 (1.5, 2.6) & 7.0 (4.6, 9.5) & 9.7 (7.4, 11.9)& 2.3 (1.6, 5.7) \\
        \hline
        \multicolumn{2}{l}{\scriptsize SD: Standard Deviation} & \multicolumn{3}{r}{\scriptsize IQR: Inter-Quartile Range} \\
      \end{tabular}
%   \end{scriptsize}
%   \end{center}
  \caption{\label{tbl:demographics} Demographic and clinical characteristics of the data.}
%   \vspace{-1em}
\end{table}

The simplest baseline method for experimental comparison is a linear regression model where each of the superior (inferior) visual field values are modeled as linear functions of all the inferior (superior) RNFL thickness values, having 19968 learnable parameters. We also compare against a fully connected neural network model in which both the superior and inferior sub-networks consist of two hidden layers with 32 nodes each and ReLU activations \cite{nair2010rectified}, followed by a 26-dimensional output layer; consequently this model has 28468 learnable parameters. To investigate the effectiveness of the recursive structure of RetiNerveNet, we also compare it against a vanilla convolutional model which employs a series of 7 successive convolutional layers to model the progression outwards from the optic disc, instead of 7 recursive passes, and also does not use skip connections. This network has 27840 learnable parameters. To put things in context, the number of learnable parameters for RetiNerveNet is 18864.

\subsection*{RetiNerveNet performance}

We compare the performance of the contending methods to reconstruct the SAP visual field values based on the SDOCT RNFL thickness values. The results are presented in Table \ref{tbl:mae} in terms of the average Mean Absolute Error (MAE) over all the 52 visual field points for the test data, with overall performance (performance on all tests) being shown alongside individual performances on tests grouped based on MD values. The tests with MD values greater than -6 are considered to exhibit the early (or no) disease, test with MD values less than -6 but greater than -12 are indicative of moderate disease, while tests having MD values lower than -12 reflect advanced disease \cite{hodapp1993clinical}. An inspection of Table \ref{tbl:mae} shows that the linear regression model overall performs worse than the other contenders, particularly on the early/no and advanced tests. This is expected owing to the simplicity of this approach. The fully connected model and the vanilla convolutional model are able to improve upon the linear regression model on both the early and advanced tests, due to their ability to model non-linear dependencies between the RNFL thickness values and the SAP values. However, both of these methods lose some performance on the moderate tests. RetiNerveNet, on the other hand, outperforms all the aforementioned contenders on all groups.

\begin{table}[ht]
%   \begin{center}
%   \begin{scriptsize}
    \centering
      \begin{tabular}{|l|l|l|l|l|}
        \hline
        Method & Early/No disease & Moderate disease & Advanced disease & Overall \\
        \hline
        Linear Regression         & $4.11 \; (0.01)$ & $6.36 \; (0.03)$ & $13.34 \; (0.04)$ & $5.54 \; (0.01)$\\
        \hline
        Fully connected model     & $3.38 \; (0.01)$ & $6.47 \; (0.03)$ & $12.65 \; (0.04)$ & $5.00 \; (0.01)$\\
        \hline
        Convolutional model       & $3.32 \; (0.01)$ & $6.68 \; (0.03)$ & $11.97 \; (0.04)$ & $4.90 \; (0.01)$\\
        \hline
        RetiNerveNet              & $3.31 \; (0.01)$ & $\mathbf{6.34} \; (0.03)$ & $11.26 \; (0.04)$ & $4.70 \; (0.01)$\\
        \hline
        RetiNN Ensemble           & $\mathbf{3.29} \; (0.01)$ & $6.43 \; (0.03)$ & $\mathbf{11.10} \; (0.04)$ & $\mathbf{4.67} \; (0.01)$\\
        \hline
        \multicolumn{5}{r}{\scriptsize Best results in \textbf{boldface}.} \\
      \end{tabular}
%   \end{scriptsize}
%   \end{center}
  \caption{\label{tbl:mae} Average MAE (with standard errors) for visual field prediction based on RNFL thickness.}
%   \vspace{-1em}
\end{table}

\subsection*{Ensemble of RetiNerveNets}

\begin{figure}[ht]
    \centering
        \subfigure[]{%
            \label{fig:tradeoff1}% label for this sub-figure
            \includegraphics[width=0.32\linewidth]{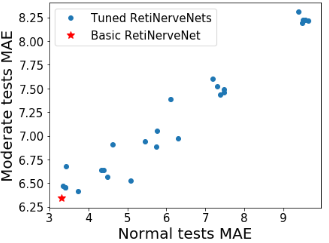}
        }%\qquad % if you want to space out the images a bit
        \subfigure[]{%
            \label{fig:tradeoff2}% label for this sub-figure
            \includegraphics[width=0.32\linewidth]{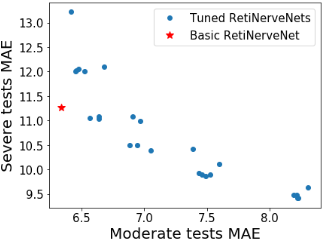}
        }
        \subfigure[]{%
            \label{fig:tradeoff3}% label for this sub-figure
            \includegraphics[width=0.305\linewidth]{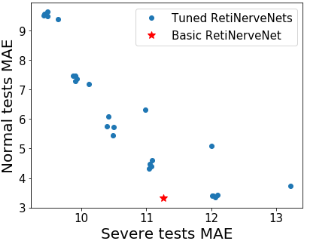}
        }
    \caption{Trade-off among the performances on the early/no, moderate, and advanced disease MD groups of tests: The performance on the advanced disease group improves when we assign higher weightage to tests from the underrepresented MD intervals, while the performance on the early/no disease and moderate disease groups worsens. Very little trade-off is observed between the early/no disease and moderate disease groups. The basic RetiNerveNet performs particularly well on the early and moderate disease tests, at the expense of the advanced disease group.}\label{fig:tradeoff}
\end{figure}

As there is imbalance in the number of tests across MD values, na{\"i}vely training the RetiNerveNet results in uneven performance across the early (or no) disease, moderate disease, and advanced disease groups. Therefore, to compensate for this imbalance, we make two modifications to the basic RetiNerveNet architecture. Firstly, we incorporate an additional output which simultaneously predicts the MD value alongside the age-corrected SAP outcomes. Further, we divide the data into 4 intervals based on the MD values, viz. (MD > -6dB), (-6dB >= MD > -16dB), (-16dB >= MD >-26dB), and (-26dB > MD), as shown in Figure \ref{fig:hist}. The bulk of the data lies within the first interval, with the amount of data progressively diminishing through the latter intervals. We then also introduce a tunable loss for training the network, which can assign higher weightage to tests from the underrepresented MD groups, and can also controls the relative importance of the SAP outcome estimation and the MD estimation. We therefore obtain multiple variants of RetiNerveNet, depending on the parameters of the loss used to train the network. The effect of tuning the loss is shown in Figure \ref{fig:tradeoff}. It is evident from Figure \ref{fig:tradeoff} that there is a clear trade-off between the performance on the advanced tests and the early and moderate tests. Consequently, some variants work better on the advanced disease group, but perform worse on the early and moderate groups, and vice-versa. The basic RetiNerveNet is observed to perform very well on the early and moderate disease tests, at the expense of the advanced disease cases. We therefore create an ensemble of all the variants of RetiNerveNet. The variant of RetiNerveNet having the lowest error for MD estimation averaged over groups is used to determine the probable group that a query RNFL thickness vector may belong to. Thereafter, the variant having the best performance (on the validation data) for the likely group is used to estimate the SAP outcomes. The performance of this ensemble is also presented in Table \ref{tbl:mae}. The ensemble outperforms all other methods in terms of overall performance as well as on the early and advanced disease cases, while it performs worse than linear regression and basic RetiNerveNet on the moderate disease tests.

\subsection*{Sectoral Averages and Mean Deviation performance}

While RetiNerveNet is trained to predict the individual age-corrected SAP values (and the MD value), we also investigate the ability of RetiNerveNet to estimate the sectoral average deviations, which are averages of the age-corrected SAP values for the different sectors of the visual field \cite{garway2000mapping}. The MAEs for estimating the sectoral average deviations and the MD values for all the models are presented in Table \ref{tbl:sectoral1}. The neural models perform better than the linear regression baseline across all sectors. Among the standalone neural models, RetiNerveNet is observed to perform the best on all the sectors as well as for MD estimation. The ensemble of RetiNerveNets outperforms the standalone methods except basic RetiNerveNet, on all sectors as well as for MD estimation. Except for the superior and superior-nasal sectors, where it performs comparatively, the ensemble also outperforms basic RetiNerveNet on the other sectors. It is also worth noting that, for sectoral average estimation, the ensemble model achieves the lowest MAE on the critical central sector.

\begin{table}[ht]
%   \begin{center}
%   \begin{tiny}
    \centering
      \begin{tabular}{|l|l|l|l|l|l|l|l|}
        \hline
        Method & Central & Temporal & Inferior & Inferior & Superior & Superior & MD \\
               &         &          &          & Nasal    &          & Nasal    &    \\
        \hline
        Linear Regression     & $4.99 \; (0.01)$ & $5.46 \; (0.01)$ & $5.67 \; (0.01)$ & $5.29 \; (0.01)$ & $5.87 \; (0.01)$ & $5.59 \; (0.01)$ & $4.31 \; (0.01)$ \\
        \hline
        Fully conn. model     & $4.42 \; (0.01)$ & $5.06 \; (0.01)$ & $5.39 \; (0.01)$ & $4.95 \; (0.01)$ & $5.33 \; (0.01)$ & $4.72 \; (0.01)$ & $3.75 \; (0.01)$ \\
        \hline
        Convolutional model   & $4.41 \; (0.01)$ & $5.28 \; (0.01)$ & $5.06 \; (0.01)$ & $4.49 \; (0.01)$ & $5.47 \; (0.01)$ & $4.80 \; (0.01)$ & $3.63 \; (0.01)$ \\
        \hline
        RetiNerveNet          & $4.40 \; (0.01)$ & $4.98 \; (0.01)$ & $4.87 \; (0.01)$ & $4.30 \; (0.01)$ & $\mathbf{5.26} \; (0.01)$ & $\mathbf{4.57} \; (0.01)$ & $3.51 \; (0.01)$ \\
        \hline
        RetiNN Ensemble       & $\mathbf{4.37} \; (0.01)$ & $\mathbf{4.97} \; (0.01)$ & $\mathbf{4.83} \; (0.01)$ & $\mathbf{4.27} \; (0.01)$ & $5.27 \; (0.01)$ & $\mathbf{4.57} \; (0.01)$ & $\mathbf{3.49} \; (0.01)$ \\
        \hline
        \multicolumn{8}{r}{\scriptsize Best results in \textbf{boldface}.} \\
      \end{tabular}
%   \end{tiny}
%   \end{center}
  \caption{\label{tbl:sectoral1} Average MAE (with standard errors) for Sectoral Averages and Mean Deviation estimation.}
\end{table}

We also compare the R$^2$ values for the different sectors for RetiNerveNet and the ensemble of RetiNerveNets with those reported recently for different deep learning methods by Christopher \emph{et al.} \cite{christopher2020deep} in Table \ref{tbl:sectoral2}. While being mindful of the fact that the quoted values were found in a study that employed a dataset different from our own, it is interesting to note that the proposed family of methods either performs much better or competitively on a majority of the sectors (viz. the central, temporal, inferior, and superior sectors), compared to more complex and rich modalities. Moreover, our methods perform better across all sectors in comparison to the quoted results using the RNFL thickness modality.

\begin{table}[ht]
    \centering
%   \begin{center}
%   \begin{tiny}
      \begin{tabular}{|l|l|l|l|l|l|l|}
        \hline
        Method & Central & Temporal & Inferior & Inferior & Superior & Superior \\
               &         &          &          & Nasal    &          & Nasal    \\
        \hline
        $^{\dagger}$ RNFL Thickness Map        & $0.08$ & $0.11$ & $0.06$ & $0.45$ & $0.31$ & $0.52$ \\
        \hline
        $^{\dagger}$ RNFL Enface Image         & $0.09$ & $0.12$ & $0.26$ & $\mathbf{0.60}$ & $\mathbf{0.35}$ & $\mathbf{0.67}$ \\
        \hline
        $^{\dagger}$ CSLO* Image               & $0.15$ & $0.08$ & $0.22$ & $0.10$ & $0.19$ & $0.26$ \\
        \hline
        $^{\dagger}$ Mean RNFL Thickness       & $0.07$ & $0.02$ & $0.01$ & $0.28$ & $0.14$ & $0.28$ \\
        \hline
        $^{\dagger}$ RNFL Thickness            & $0.07$ & $0.01$ & $0.02$ & $0.36$ & $0.17$ & $0.31$ \\
        \hline
        RetiNerveNet                           & $0.38$ & $0.24$ & $0.40$ & $0.48$ & $0.32$ & $0.47$ \\
        \hline
        RetiNN Ensemble                        & $\mathbf{0.39}$ & $\mathbf{0.25}$ & $\mathbf{0.41}$ & $0.49$ & $0.32$ & $0.48$ \\
        \hline
        \multicolumn{5}{l}{\scriptsize *CSLO: Confocal Scanning Laser Ophthalmoscopy} & \multicolumn{2}{r}{\scriptsize Best results in \textbf{boldface}.} \\
        \multicolumn{5}{l}{\scriptsize $\dagger$ Quoted from Christopher \emph{et al.} \cite{christopher2020deep}} & \multicolumn{2}{r}{} \\
      \end{tabular}
%   \end{tiny}
%   \end{center}
  \caption{\label{tbl:sectoral2} R$^2$ values for Sectoral Mean Pattern Deviation estimation.}
\end{table}

\section*{Discussion}

We propose RetiNerveNet, a deep fully convolutional neural architecture for obtaining estimates of SAP visual field values based on RNFL thickness values obtained from the more objective SDOCT tests. Unlike existing works with similar aim \cite{zhu2010predicting, zhu2011quantifying, guo2017optical, maetschke2019inference, christopher2020deep, mariottoni2020artificial, park2020deep}, we postulate that building our network to mimic the arcuate structure of the axons of the retinal ganglion cells can help improve performance for this task. The fact that the proposed architecture performs better than a number of baselines in Table \ref{tbl:mae} seems to corroborate our hypothesis. Moreover, RetiNerveNet also performs better than its vanilla fully convolutional counterpart, suggesting that it may be possible to model seemingly complex biological structures, like that of the retinal axon fibers, as recursive passes through a single function (the RPL in our case).

We also show that an ensemble of RetiNerveNets can further improve performance on early/no and advanced disease groups, beyond that of the basic RetiNerveNet. The improvement achieved by the ensemble is the result of using a tunable loss function which can be used to focus the attention on the underrepresented subsets of tests (i.e, the moderate and advanced MD groups in our case). This results in improved performance on different subsets (at the expense of the other subsets) by the different models. Combining all of these variants into an ensemble help us leverage this improved performance for each of the subsets, resulting in overall improvement. It is also reassuring to observe that the ensemble achieved the lower MAE on the central points of the visual field, which are known to be very important from a clinical point of view \cite{abe2016impact}. Moreover, comparison with the results reported by Christopher \emph{et al.} \cite{christopher2020deep} shows that our results are characterized by more uniform performance on all the sectors of the visual field. The results from the former article seem to imply that rich modalities like an enface image or an RNFL thickness map are required to achieve good performance on this task. Our results, on the contrary, show that much of the information needed to estimate visual function may be extracted from the RFNL thickness vector, using a more sophisticated model like RetiNerveNet.

Since we undertake supervised deep learning in this study, the main limitations of our work are that the predictions from our models can only be as good as visual field estimates obtained from SAP, which are known to be noisy. Moreover, as the RNFL thickness values were obtained from the SDOCT scans using the conventional SDOCT software, errors inherent to that process \cite{mariottoni2020quantification} may also have affected our results. An interesting future direction of research may be to investigate how our proposed architecture may perform in conjunction with segmentation-free RNFL thickness extraction systems based on deep learning \cite{mariottoni2020quantification}.

\section*{Methods}

Our retrospective study is based on cross-sectional data about paired SDOCT and SAP tests, from the Duke Glaucoma Registry, a database of electronic medical and research records at the Vision, Imaging, and Performance Laboratory at Duke University. Healthy individuals as well as patients having or suspected of having glaucoma were included in the study. Diagnosis of glaucoma or suspect was based on International Classification of Diseases (ICD) codes \cite{jammal2020impact}. We excluded patients who underwent procedures such as panretinal photocoagulation or suffered from other diseases like retinal detachment, optic neuritis, proliferative diabetic retinopathy, etc. that could impact the RNFL thickness measurements from SDOCT, or could impact the SAP visual fields. Patients younger than 18 years were also excluded. This study was approved by the Duke University Institutional Review Board. Due to the retrospective nature of the work, a waiver of informed consent was granted. All methods adhered to the tenets of the Declaration of Helsinki for research involving human participants, and the study was in accordance with the regulations of the Health Insurance Portability and Accountability Act.

\subsection*{Standard Automated Perimetry}

The SAP visual field tests were performed using the 24-2 Swedish Interactive Threshold Algorithm (SITA) protocol (Carl Zeiss Meditec, Inc., Dublin, CA; \cite{bengtsson1997new}). We excluded unreliable tests with more than 33\% fixation losses or more than 15\% false-positive errors. Out of the 54 visual field values obtained using the 24-2, , the two values corresponding to points around the blind spot were removed, resulting in 52 sensitivity threshold values being obtained for each SAP test. The locations in the superior hemiretina are responsible for inferior vision (and vice versa). The visual fields were also corrected for age to obtain the age-corrected threshold (a.k.a. total deviation) values, which were to be predicted, for the 52 locations.

\subsection*{Spectral-Domain Optical Coherence Tomography}

RNFL thickness was collected from peripapillary RNFL scans, acquired using the Spectralis SDOCT (Software version 5.4.7.0, Heidelberg Engineering, GmbH, Dossenheim, Germany). RNFL thickness values at 768 evenly spaced points on a circle having a diameter of 3.45 mm, positioned around the center of Bruch’s membrane opening (the opening of the optic disc), was obtained directly form the SDOCT software. Tests having a quality score lower than 15 were excluded, according to the manufacturer recommendations. 

\subsection*{RetiNerveNet}

\begin{figure}[ht]
  \centering 
  \includegraphics[width=0.85\textwidth]{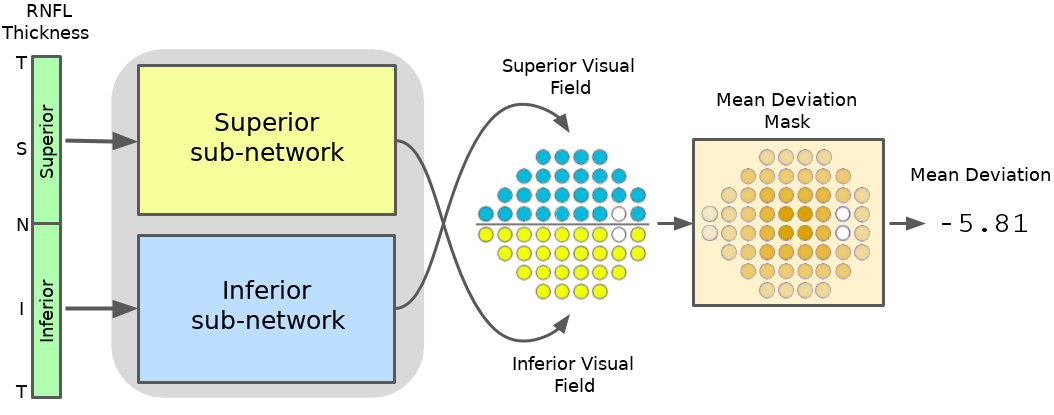} 
  \caption{Overview of the RetiNerveNet structure: A 768-dimensional RNFL thickness vector obtained from the SDOCT test, arranged in the Temporal-Superior-Nasal-Inferior-Temporal (TSNIT) order, is split into two halves having size 384. As the retinal ganglion fibers in the eye do not cross the horizontal raphe separating the superior and inferior hemiretinae, we use separate sub-networks to estimate the superior and inferior visual fields. The superior (inferior) half of the RNFL thickness values proceeds through the superior (inferior) sub-network of the RetiNerveNet to yield an estimate of the inferior (superior) half of the visual field. For the multi-task version of RetiNerveNet, a weight mask is applied to the estimated visual field to obtain a scalar estimate of Mean Deviation. %The inferior half of the RNFL thickness, on the other hand, proceeds through the inferior sub-network to produce an estimate of the superior visual field.
  }
  \label{fig:overview} 
\end{figure}

The RNFL is comprised of the axons of the retinal ganglion cells, which converge to the optic disc and emerge from the eye as the optic nerve \cite{weinreb2004primary}. Our task is the regression problem of estimating the 52 visual field values (excluding the 2 points around the blind spot) from the SAP test as a function of 768 RNFL thickness values obtained from the SDOCT test. To be able to do this, we must essentially trace-back the path along which the retinal axon fibers converge to the optic disc (see Figure \ref{fig:rnfl}). The axons from retinal ganglion cells located in the superior half of the retina (superior hemiretina) do not cross to the inferior half (inferior hemiretina), and vice versa. Axons just superior to the raphe (which separates the two hemiretinae) are directed superiorly. In contrast, neighboring fiber just inferior to the raphe abruptly take the opposite direction. Therefore, our proposed RetiNerveNet architecture consists of two separate, similarly structured sub-networks, viz. the superior and inferior sub-networks for respectively processing the superior and inferior halves of the RNFL thickness vector, as illustrated in Figure \ref{fig:overview}. 

% \vspace{-3mm}

Within each hemiretina, the axons converge to the optic disc in a characteristic arcuate pattern. As can be seen from Figure \ref{fig:rnfl}, the axon fibers from the ganglion cells located in the temporal side of the retina do not cross the center of the fovea in their trajectory to the optic disc, but rather deviate from it radially. Additionally, it is also worth noting that the axons from the distal cells follow a path similar to the more proximal fibers but are further away from the fovea. Hence, the fibers do not cross over each other and determine a unique direction at any point on the retina. The RNFL thickness values obtained from the SDOCT test represents information at a much finer resolution than is captured by the SAP test. Further, in keeping with the fact that several axons converge into the same part of the optic disc, any defect in the optic disc is likely to impact a large number of spatially proximal ganglion cells. Consequently, each visual field point may depend on multiple RNFL thickness values in the vicinity of the point on the optic disc to which the axon in question converges. On the other hand, due to the property that the ganglion axons in the RNFL do not cross each other, ganglion cells (and therefore visual field points) that are sufficiently far apart should have low correlation. The only correlations between such distant visual field points is expected to come from factors having a global effect, such as test conditions or patient cooperation. Convolutional networks, in conjunction with pooling layers (like max pooling \cite{yamaguchi1990neural}), are effective for distilling coarse information from finer details while preserving spacial correlation \cite{lecun1998gradient}. Hence, we refrain from using fully connected layers (which could potentially result in spatially distant visual field points having correlated values) and instead constrain RetiNerveNet to entirely consist of 1D convolutional layers (along with max pooling). SAP measures the visual function at 52 locations on the retina, with the 26 locations in the superior hemiretina being responsible for inferior vision (and vice versa) due to lateral inversion in the eye. SDOCT, on the other hand, scans the nerve tissue along a circle around the optic disc.

\begin{figure}[ht]
  \centering 
  \includegraphics[width=1.02\textwidth]{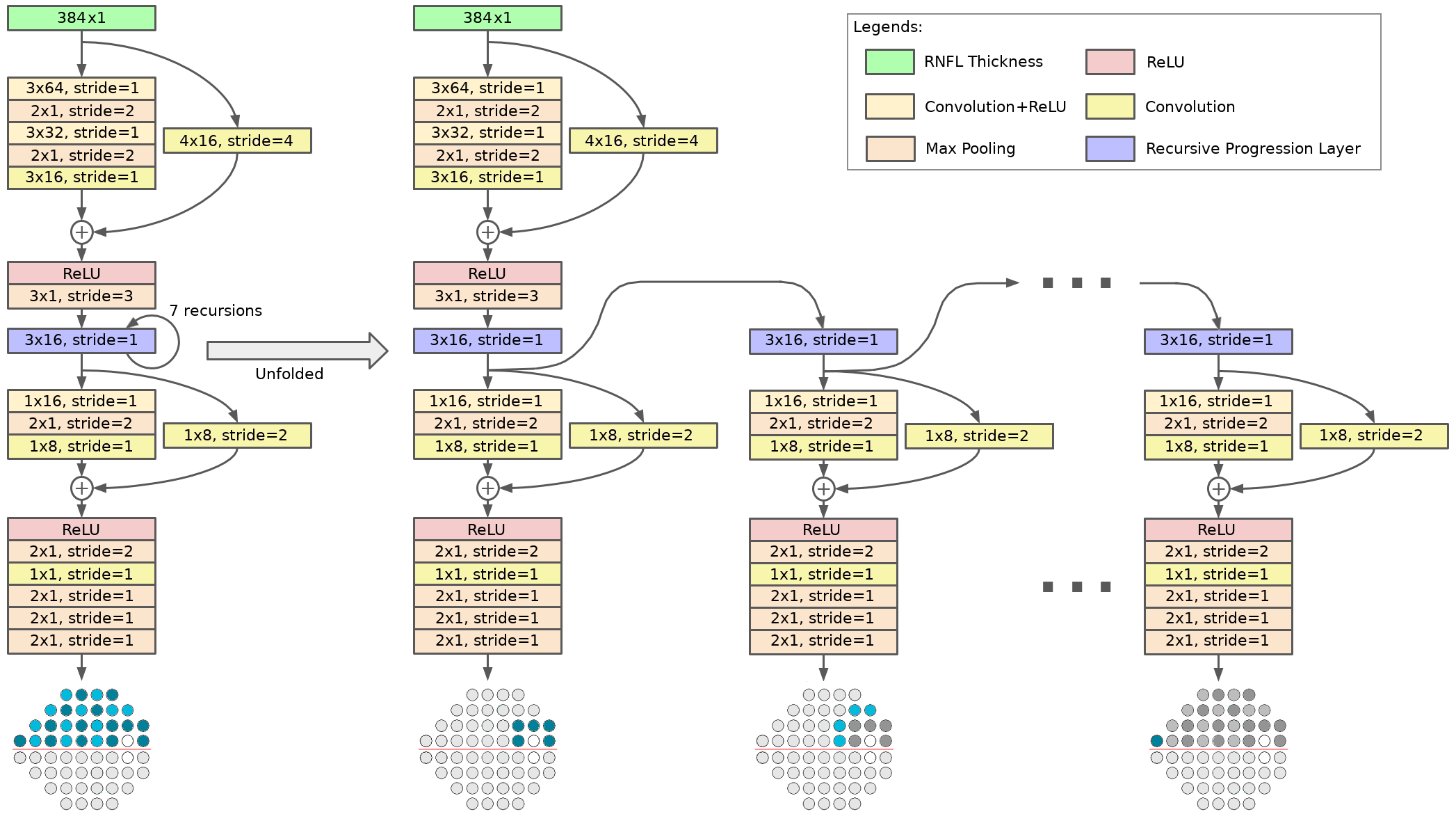} 
  \caption{%\textcolor{blue}{1. I would move the text here to the main body. 2. I would add another recurrent layer (number 3) in the unfolded version, maybe instead of the last one and maybe instead of the folded version. 3. I would change the way you color in blue the SAP components such that in layer n, I would color all the SAP from layers 1,2,..n. You can color SAPs of layer 1,2,...,n-1 with a slightly different color than layer n} 
  Details of the inferior sub-network of RetiNerveNet (best viewed in color): The sub-network has four main blocks. The first block of convolutional and max pooling layers, augmented by a convolutional skip connection. The second block only consists of the Recursive Progression Layer (RPL) which recursively models the outward spatial movement from the optic disc. None of the blocks except the RPL, correspond to any spatial movement along the retina. The next block consists of convolutional and max pooling layers, augmented by another convolutional skip connection. The final block consists of a single convolutional layer which combines the different representations obtained from the previous block into a scalar value. The subsequent max pooling layers bring down the dimension of the output to the intended size (5-dimensional vector). All convolutional layers are equipped with ReLU activations, except both the skip connections and the final convolutional layers of the corresponding blocks. The network respectively uses the first 5, 4, 4, 4, 4, 4, and 1 dimensions of the outputs after 7 successive recursive passes to obtain an estimate of the entire superior half of the visual field (consisting of 26 points). The visual field points estimated after successive passes are differentiated using contrasting shades. 
% Finally, the ordinal regression layer maps the scalar value obtained for each visual field point to a 61-dimensional vector.
  % Details of the inferior sub-network of RetiNerveNet (best viewed in color): The sub-network has four main blocks. The first, third and fourth blocks consist of convolutional and max pooling layers; the former two being augmented by convolutional skip connections. The second block only consists of the Recursive Progression Layer (RPL) which recursively models the outward spatial movement along the retina from the optic disc. All convolutional layers are equipped with ReLU activations, except both the skip connections and the final convolutional layers of the corresponding blocks. While the output is always 5-dimensional, the network respectively uses the first 5, 4, 4, 4, 4, 4, and 1 dimensions of the outputs after 7 successive recursive passes to obtain an estimate of the entire superior half of the visual field (consisting of 26 points). The visual field points estimated after successive passes are differentiated using contrasting shades. Finally, the ordinal regression layer maps the scalar value obtained for each visual field point to a 61-dimensional vector.
  }
  \label{fig:details} 
\end{figure}

The structure of the inferior sub-network is further detailed in Figure \ref{fig:details}. Each sub-network consists of four main blocks. The first block is composed of a series of convolutional and max pooling layers, and is meant to extract coarse structural information from the fine-grained RNFL thickness values. The second block consists of the RPL. Each recursive pass through the RPL is meant to model further spatial progression through the retina outwards from the optic disc. The third block, akin to the first, also consists of convolutional and max pooling layers. This block is meant to extract rich information about the outputs from the RPL by projecting onto different subspaces, while also further coarsening the resolution. The final block consists of a single convolutional layer followed by a series of max pooling layers. This layer combines the different representations obtained from the previous block into a single scalar value and the subsequent pooling layers reduce the resolution to the intended 5-dimensions. Deeper neural networks are known to be more difficult to train and a common solution is to introduce skip connections into the network \cite{he2016deep}. Therefore, in order to ease the training of our proposed RetiNerveNet, we incorporate convolutional skip connections with appropriate filter size and strides in the first and third blocks of RetiNerveNet. It is important to notice that the first , third, and fourth convolutional blocks are not responsible for modeling any spatial movement. Spatial movement along the retina is modeled solely by the RPL.

\subsubsection*{Recursive Progression Layer}%\label{sec:rpl}
We trace-back the arcuate convergence of the axons fibers as multiple recursive passes through a single convolutional layer with linear activation. In other words, the first pass through the RPL is meant to map from the SDOCT circle to the region of the retina corresponding to the first 5 visual field points nearest to the blind spot. As we move further outwards, the additional increment in curvature can be captured by applying the same transformation to the output obtained after the first pass. Thus, the second recursive pass is meant to map (from the first 5 visual field points) to the region corresponding to the next 4 visual field points, and so on (see Figure \ref{fig:details}). Due to the use of convolution to model this progression, a small sector on the SDOCT circle can affect a larger number of visual field points as we move further away from the disc. Since RNFL axons which are wide apart in the retina are known to converge into the same region on the optic disc, this divergent property of the recursive layer is desirable for the intended trace-back. Formally, starting with $\mathbf{r}^{(0)} = f_{\theta_1}(\mathbf{x})$, $f_{\theta_1}(.)$ being the first block of RetiNerveNet and $\mathbf{x}$ being the input RNFL thickness vector, we have
\begin{equation}
    \mathbf{r}^{(t)} = RPL_{\theta_r}(\mathbf{r}^{(t-1)}),
\end{equation}
where $\mathbf{r}^{(t)}$ ($t \leq 7$) denotes the output after the $t$-th pass through the RPL.

While the output after the four blocks of the superior (or inferior) sub-network is always 5-dimensional, only the first 5, 4, 4, 4, 4, 4, and 1 dimensions of the output for the respective passes are deemed to be points in the visual field, as the rest are deemed to have progressed outside the region tested by SAP. Therefore, after 7 recursive passes through the RPL (and the subsequent part of the sub-network), the entire inferior (or superior) half of the visual field (consisting of 26 points) is estimated.

% Formally, we intend to estimate the vector $\mathbf{y} \in \mathbb{R}^{52}$ of SAP outcomes given the corresponding vector $\mathbf{x} \in \mathbb{R}^{768}$ of RNFL thickness values obtained from the SDOCT test. Consider also that we have access to a training dataset $\mathcal{D} = \{(\mathbf{x}_i, \mathbf{y}_i) | \mathbf{x}_i \in \mathbb{R}^{768}, \mathbf{y}_i \in \mathbb{R}^{52}, i \in \{1,\cdots,N\}\}$, of $N$ paired RNFL thickness-SAP visual field values. 

\subsubsection*{Loss Function}\label{sec:loss}

A combination of two losses is used to train the RetiNerveNet. The first loss is calculated on estimated individual visual field points, while the second loss is calculated on the estimated MD value. The loss on the estimated visual field values is
\begin{equation}
    \mathcal{L}_{VF} = \sum_{i=1}^N \lambda_i \sum_{j=1}^{52} \rho_j (y_{i,j} - F_{\theta}^{(t_j)}(\mathbf{x}_i))^2,
\end{equation}
where $N$ is the number of training SDOCT-SAP pairs, $\lambda_i$ is the cost associate with the $i$-th SDOCT-SAP pair, $\rho_j$ is the cost associated with the $j$-th SAP location, $y_{i,j}$ is the true age-corrected threshold value at the $j$-th SAP location for the $i$-th SDOCT-SAP pair, $F_{\theta}^{(t)}(\mathbf{x}_i) = (h_{\theta_3}\; \circ \;\; g_{\theta_2})(\mathbf{r}^{(t)}_i)$ is the output of RetiNerveNet after the $t$-th recursive pass ($g_{\theta_2}$ and $h_{\theta_3}$ respectively being the third and fourth RetiNerveNet blocks), and $t_j$ is the recursive pass corresponding to the $j$-th visual field location. Further, the loss on the estimated MD values is 
\begin{equation}
    \mathcal{L}_{MD} = \sum_{i=1}^N \lambda_i (z_i - M_{\theta}(\mathbf{x}_i))^2,
\end{equation}
where $z_i$ is the true MD value for the $i$-th SDOCT-SAP pair, and $M_{\theta}(\mathbf{x}_i)$ is the corresponding estimate based on the output $F_{\theta}^{(t)}(\mathbf{x}_i)$ from the network. Hence, the total loss used for training the RetiNerveNet is
\begin{equation}
    \mathcal{L} = (1 - \beta)\mathcal{L}_{VF} + \beta\mathcal{L}_{MD},
\end{equation}
where $\beta$ is a tunable parameter controlling the trade-off between the two losses. The cost $\lambda_i$ assigned to the $i$-th SDOCT-SAP pair on the basis of which of the 4 MD value intervals (see Figure \ref{fig:hist}) it belongs to, is given by
\begin{equation}
    \lambda_i = (1 - \alpha) \frac{1}{N} + \alpha \frac{1}{4 N_i}, %3 bins
\end{equation}
where $N_i$ is the number of training SDOCT-SAP pairs available from the interval to which the $i$-th pair belongs, and $\alpha$ is the hyperparameter controlling the relative importance of the underrepresented intervals. The cost $\rho_j$ associate with the $j$-th SAP location is given by
\begin{equation}
    \rho_j = \frac{\exp(-d_j^2/2\gamma^2)}{\sum_{j=1}^{52} \exp(-d_j^2/2\gamma^2)},
    \label{eq:weights}
\end{equation}
where $d_j$ is the Euclidean distance of the $j$-th SAP location w.r.t. the center of the SAP visual field (see Figure \ref{fig:rnfl2}), with unit distance between adjacent SAP locations, and $\gamma$ is a tunable parameter determining how much more important the central points of the visual field are compared to the peripheral points.

\subsubsection*{Experiment Setup and Ensemble}

The RetiNerveNet has three tunable parameters $\alpha$, $\beta$, and $\gamma$. Setting $\alpha=0$ and $\beta=0$ gives us the basic RetiNerveNet. We varied $\gamma$ in the set $\{0.5, 5, 50\}$ and found $\gamma=5$ to yield the lowest MAE. Therefore, we set $\gamma=5$ for all other experiments. We varied $(\alpha, \beta)$ in $G \times G$, $G=\{0.01, 0.25, 0.5, 0.75, 0.99\}$, to obtain different variants of RetiNerveNet. We conduct 5 independent runs of RetiNerveNet with each setting for a maximum of 2000 epochs. Early stopping as well as the selection of the best among the early stopped models from the independent runs is undertaken based on lowest validation loss. We further combine all the obtained variants of RetiNerveNet (except the basic variant) into an ensemble with the following steps:
\begin{enumerate}
    \item \label{algstep:bestassign} We choose the variant that minimizes the average MAE for MD estimation across the early, moderate, and advanced groups of tests in the validation data.
    \item \label{algstep:assign} We then assign each SDOCT-SAP pair in the test data to a particular group, using the variant of RetiNerveNet chosen in Step \ref{algstep:bestassign}.
    \item \label{algstep:bestinfer} Thereafter, for each group, we find the variant having minimum average MAE on the validation data.
    \item Finally, for all the SDOCT-SAP pairs assigned to a particular group in Step \ref{algstep:assign}, we infer the SAP visual field using the best RetiNerveNet variant, for the group in question, found in Step \ref{algstep:bestinfer}.
\end{enumerate}

\bibliography{refs2}

% \noindent LaTeX formats citations and references automatically using the bibliography records in your .bib file, which you can edit via the project menu. Use the cite command for an inline citation.

% For data citations of datasets uploaded to e.g. \emph{figshare}, please use the \verb|howpublished| option in the bib entry to specify the platform and the link, as in the \verb|Hao:gidmaps:2014| example in the sample bibliography file.

% \section*{Acknowledgements (not compulsory)}

% Acknowledgements should be brief, and should not include thanks to anonymous referees and editors, or effusive comments. Grant or contribution numbers may be acknowledged.

\section*{Author contributions statement}

S.D., E.B.M. and D.D. conceived and designed the study described here. S.D. conducted the experiments. E.B.M. and A.A.J. collected and managed data. L.C. and F.A.M. advised in study design, data analysis methods, and drafting the manuscript. All authors reviewed the manuscript.

% \section*{Additional information}

% \textbf{Competing interests}: The author(s) declare no competing interests.

% To include, in this order: \textbf{Accession codes} (where applicable); \textbf{Competing interests} (mandatory statement). 

% The corresponding author is responsible for submitting a \href{http://www.nature.com/srep/policies/index.html#competing}{competing interests statement} on behalf of all authors of the paper. This statement must be included in the submitted article file.

% \begin{figure}[ht]
% \centering
% \includegraphics[width=\linewidth]{stream}
% \caption{Legend (350 words max). Example legend text.}
% \label{fig:stream}
% \end{figure}

% \begin{table}[ht]
% \centering
% \begin{tabular}{|l|l|l|}
% \hline
% Condition & n & p \\
% \hline
% A & 5 & 0.1 \\
% \hline
% B & 10 & 0.01 \\
% \hline
% \end{tabular}
% \caption{\label{tab:example}Legend (350 words max). Example legend text.}
% \end{table}

% Figures and tables can be referenced in LaTeX using the ref command, e.g. Figure \ref{fig:stream} and Table \ref{tab:example}.

\end{document}